\documentclass[runningheads,a4paper]{llncs}

\usepackage{amssymb}
\usepackage{multirow}
\usepackage[noend]{algorithmic}

\newcommand{\keywords}[1]{\par\addvspace\baselineskip
\noindent\keywordname\enspace\ignorespaces#1}

\usepackage[final]{graphicx}
\usepackage[noend]{algorithmic}
\usepackage{algorithm}
\usepackage{latexsym}

\usepackage{amsmath}
\newcommand{\argmax}{\operatornamewithlimits{arg\,max}} 

\begin{document}
\title{Feature Construction for\\ Relational Sequence Learning}

\author{Nicola Di  Mauro \and Teresa M.A. Basile  \and Stefano Ferilli
  \and Floriana Esposito} 
\institute{Universit\`a degli Studi di Bari, Dipartimento di Informatica,
70125 Bari, Italy\\
\{ndm,basile,ferilli,esposito@di.uniba.it\}
}
\maketitle

\begin{abstract}  We tackle  the problem  of  multi-class relational  sequence learning  using
  relevant  patterns discovered  from  a set  of labelled  sequences.   To deal  with this  problem,
  firstly each relational sequence is mapped into a feature vector using the result of a feature construction
  method.
 Since, the efficacy of sequence learning algorithms strongly depends on the features used
  to  represent the sequences,  the second  step is  to find  an optimal  subset of  the constructed
  features leading  to high  classification accuracy.  This feature selection  task has  been solved
  adopting a wrapper approach  that uses a stochastic local search algorithm  embedding a na\"ive Bayes
  classifier.
  The performance of the proposed method applied to a real-world dataset shows an improvement when
  compared  to  other  established  methods,  such  as hidden  Markov  models,  Fisher  kernels  and
  conditional random fields for relational sequences.

\keywords{Relational  Sequence Learning,  Feature Construction/Selection,  Stochastic  Local Search,
  Statistical Relational Learning.} 
\end{abstract}

\section{Introduction}
\label{sec:Introduction}

Sequential reasoning is a fundamental task of  intelligence. Indeed, sequential data may be found in
a lot of contexts of the every day life.  From a computer science point of view, sequential data may
be found  in many applications  such as video  understanding, planning, computational  biology, user
modelling, speech recognition, etc. 
The sequences  are the simplest  form of structured  patterns and different methodologies  have been
proposed to face the problem of sequential pattern mining, firstly introduced in~\cite{agrawal95}, with the aim of capturing the existent maximal frequent sequences in
a given database.  One  of the many problems investigated concerns assigning  labels to sequences of
objects.  
However,  some  environments involve  very  complex components  and  features.  Thus, the  classical
existing data mining approaches,  that look for patterns in a single  data table, have been extended
to the  multi-relational data  mining approaches  that look for  patterns involving  multiple tables
(relations)  from a  relational  database. This  has  led to  the exploitation  of  a more  powerful
knowledge representation formalism as first-order logic.  

Indeed, sequential learning techniques may be classified by the language they adopt to describe
sequences.  On  the one hand we  find algorithms adopting  a propositional language, such  as hidden
Markov models (HMMs)~\cite{rabiner86}, allowing both a simple model representation and an efficient algorithm; on the
other hand  probabilistic relational  systems are  able to elegantly  handle complex  and structured
descriptions where, on the contrary, an  atomic representation could make the problem intractable to
propositional sequence learning techniques. The aim of  this paper is to propose a new probabilistic
algorithm for relational sequence learning~\cite{kersting:rslchapter08}.

A way to  tackle the task of  learning discriminant functions in relational  learning corresponds to
reformulate  the   problem  into  an  attribute-value   form  and  then   applying  a  propositional
learner~\cite{kramer01a}.  The  reformulation  process  may  be  obtained  adopting  a  \emph{feature
  construction} method, such as  mining frequent patterns that can then be  successfully used as new
Boolean features~\cite{Dehaspe98,King01,Kramer01}. 
Since,  the efficacy  of  learning  algorithms strongly  depends  on the  features used  to
represent the sequences, a \emph{feature selection} task should be very useful.
The  aim of  feature
  selection is to find an optimal subset of the input features leading to high 
classification performance,  or, more generally, to carry  out the classification task  in a optimum
way.  However, the  search for  a variable  subset  is a  NP-hard problem.   Therefore, the  optimal
solution cannot  be guaranteed to  be acquired  except when performing  an exhaustive search  in the
solution space. The  use of a \emph{stochastic local  search} procedure allows one to  obtain good solutions
without having to explore the whole solution space. 
Algorithms  for  feature  selection  can  be   divided  into  two  categories:  wrapper  and  filter
methods~\cite{guyon03}. When the feature selection algorithm embeds a classifier and selects
subsets of features guided by their predictive power predicted by the classifier, it is 
using a wrapper approach.  The filter approach selects the features adopting a preprocessing
step using heuristics based on the intrinsic characteristic of the data and ignoring the learner. 

In  this  paper  we propose  a  new  algorithm,  named \textsf{Lynx}\footnote{\textsf{LYNX}  is  public
  available  at \texttt{http://www.di.uniba.it/$\sim$ndm/lynx/}.}  for relational  sequence learning
that in the first step it adopts
a classical feature  construction approach.  As we will  see in the following, here  the features are
not considered as Boolean but we are able to associate a probability to each constructed feature. 
In the second step, the system adopts a wrapper feature selection approach, that uses
a stochastic local  search (non-exhaustive) procedure, embedding a na\"ive  Bayes classifier to select
an optimal subset of the constructed features.  In particular, the optimal subset of
patterns is searched using a Greedy Randomised  Search Procedure (GRASP) and the search is guided by
the predictive power of the selected subset computed using a na\"ive Bayes approach. 

Hence  the focus  of  the  paper is  on  combining probabilistic  feature  construction and  feature
selection  for relational  sequence learning.  The aim  is  to show  that the  proposed approach  is
comparable   to  other  purposely   designed  probabilistic   approaches  for   relational  sequence
learning. 

The outline of the paper is  as follows. After discussing related work in Section~\ref{sec:relwork},
we present the \textsf{Lynx} algorithm in Section~\ref{sec:Lynx}. In particular we will briefly
present  the description  language, followed  by  the description  of the  feature construction  and
feature selection proposed methods. Before concluding the paper in Section~\ref{sec:conclusions}, we
experimentally evaluate \textsf{Lynx} on a real-world dataset.

\section{Related Work}
\label{sec:relwork}

As already pointed out, the  problem of sequential pattern mining is a central one  in a lot of data
mining applications and many  efforts have been done in order to  propose purposely designed methods
to face it. Most of the works have  been restricted to propositional patterns, that is, patterns not
involving first order predicates. One of the early domains that highlights the need to describe with
structural information the  sequences was the bioinformatics. Thus, the need  to represent many real
world domains  with structured data sequences  became more unceasing, and  consequently many efforts
have been  done to extend  existing or propose  new methods to  manage sequential patterns  in which
first order predicates are involved. 
Related works may be divided into two categories.  In the first category there are work belonging to
the  Inductive Logic  Programming area~\cite{muggleton94},  that  reformulate the  initial relational  problem into  an
attribute-value  form,  by using  frequent  patterns  as new  Boolean  features,  and then  applying
propositional learners.  To the second category  belong all the systems purposely designed to tackle
the problem  of relational sequence analysis  falling into the more  specific Statistical Relational
Learning area~\cite{1296231} where probabilistic models are combined with relational learning. 

This work may be correlated to that in \cite{Kramer01}, where the authors presented one
of the first Inductive Logic Programming  feature construction method.  They firstly construct a set
of features adopting a declarative language to  constraint the search space and to find discriminant
features.   Then, these  features are  used to  learn a  classification model  with  a propositional
learner. 

In \cite{lee04} are presented  a logic language, SeqLog, for mining sequences  of logical atoms, and
the inductive mining system MineSeqLog, that  combines principles of the level-wise search algorithm
with the version space  in order to find all patterns that satisfy a  constraint by using an optimal
refinement  operator for  SeqLog.  SeqLog is  a  logic representational  framework  that adopts  two
operators  to represent the  sequences: one  to indicate  that an  atom is  the direct  successor of
another and the other to say that an atom occurs somewhere after another. Furthermore, based on this
language, the notion of subsumption, entailment and a fix point semantic are given. 

These work even if may be correlated to  our work, they tackle into account the feature construction
problem  only. Here, however  we combine  a feature  construction process  with a  feature selection
algorithm maximising the  predictive accuracy of a probabilistic model. Systems  very similar to our
approach are those that combine a probabilistic models with a relational description such as logical
hidden  Markov  models  (LoHHMs)~\cite{Kersting06},   Fisher  kernels  for  logical  sequences~\cite{KerstingG04},  and
relational conditional random  fields~\cite{Gutmann06} that are purposely designed  for relational sequences
learning.  

In \cite{KerstingG04} has been proposed an extension of classical Fisher kernels, working on sequences
over flat alphabets, in order to make them  able to model logical sequences, i.e., sequences over an
alphabet of  logical atoms. Fisher kernels were  developed to combine generative  models with kernel
methods,  and have shown  promising results  for the  combinations of  support vector  machines with
(logical) hidden Markov  models and Bayesian networks.  Successively,  in \cite{Kersting06} the same
authors proposed an algorithm for selecting LoHMMs from data. HMM~\cite{rabiner86} are one of the most popular methods for analysing sequential data, but they can be exploited
to handle  sequence of flat/unstructured  symbols. The proposed  logical extension~\cite{Kersting05}
overcomes such weakness by handling sequences of  structured symbols by means of a probabilistic ILP
framework. 

Finally, in~\cite{Gutmann06} an extension of conditional random fields (CRFs) to logical sequences has been
proposed. In the case  of sequence labelling task, CRFs are a better  alternative to HMMs that makes
it relatively easy to model arbitrary dependencies in the input space. CRFs are undirected graphical
models that  instead of learning a  generative model, such as  in HMMs, they  learn a discriminative
model designed  to handle non-independent input  features. In~\cite{Gutmann06},  the authors lifted
CRFs  to the  relational  case  by representing  the  potential functions  as  a  sum of  relational
regression trees learnt by a relational regression tree learner.

\section{\textsf{Lynx}: a relational pattern-based classifier}
\label{sec:Lynx}
This section firstly briefly reports the framework for mining (multi-dimensional) relational sequences
introduced in~\cite{esposito08fundamenta}  to manage patterns in which more than one dimension is taken into
account.  That framework has  been used in \textsf{Lynx} due to its  general logic formalism for representing
and mining relational  sequences. Over that framework \textsf{Lynx}  implements a probabilistic pattern-based
classifier.  In  particular, after  introducing the representation  language, the \textsf{Lynx}  system along
with its  feature construction capability, the  adopted pattern-based classification  model, and the
feature selection approach will be presented.

\subsection{The language}
As a representation language we used a first-order logic that we briefly review. 
The   first-order   \emph{alphabet}  consists   of   a  set   of   \textit{constants},   a  set   of
\textit{variables}, a  set of  \textit{function symbols}, and  a non-empty set  of \textit{predicate
  symbols}. Both function  symbols and predicate symbols have a  natural number (its \textit{arity})
assigned to it. 
A \emph{term} is a constant symbol, a variable symbols, or an $n$-ary function symbol $f$ applied to
n terms $t_1, t_2, \ldots, t_n$. 

An atom $p(t_1, \ldots, t_n)$ (or atomic formula)  is a predicate symbol $p$ of arity $n$ applied to
$n$ terms $t_i$.  
Both $l$ and its negation $\overline{l}$ are said to be \emph{literals} (resp. positive and negative
literal) whenever $l$ is an atomic formula. 

A \emph{clause}  is a  formula of the  form $\forall X_1  \ldots \forall  X_n (L_1 \vee  \ldots \vee
\overline{L}_i \vee \overline{L}_{i+1} \vee \ldots \vee \overline{L}_m)$ where each 
$L_i$ is  a literal and $X_j,  j=1,\ldots,n$, are all the  variables occurring in  the literals. The
same clause may be written as $L_1, \ldots \leftarrow L_i, \ldots L_m$.  

Clauses, literals and terms  are said to be \emph{ground} whenever they  do not contain variables. A
\emph{Datalog clause}  is a clause with  no function symbols  of non-zero arity; only  variables and
constants can be used as predicate arguments. 

A \emph{substitution} $\theta$ is defined as a set of bindings $\{X_1 \leftarrow a_1, \ldots, X_n \leftarrow a_n\}$ where $X_i, 1 \leq i \leq n$ is a variable and $a_i, 1 \leq i \leq n$ is a term. A substitution $\theta$ is applicable to an expression $e$, obtaining the expression $e\theta$, by replacing all variables $X_i$ with their corresponding terms $a_i$.

\textsf{Lynx}  includes the  multi-dimensional relational  framework,  and the  corresponding pattern  mining
algorithm, reported in~\cite{esposito08fundamenta} that here we briefly recall.

A \textit{1-dimensional relational sequence} may be defined as an ordered list of Datalog atoms
separated by the operator $<$: $l_1 < l_2 < \cdots < l_n$. 

Considering a sequence as an ordered succession of events for each dimension, fluents have been used
to indicate that an atom is true for a given event. 
For the general case of $n$-dimensional sequences, the operator $<_i$ has been introduced to
express multi-dimensional relations.  Specifically, \texttt{($e_1 <_i e_2$)} denotes  that the event
$e_2$  is the  successor  event of  $e_1$ on  the  dimension $i$.  Hence, A  \emph{multi-dimensional
  relational sequence} may be defined as a set of Datalog atoms, concerning $n$
dimensions, where  each event may be  related to another event  by means of the  $<_i$ operators, $1
\leq i \leq n$.  

In order  to represent  multi-dimensional relational patterns,  the following  dimensional operators
have been  introduced. Given a  set $\mathcal D$  of dimensions, $\forall  i \in \mathcal  D$: $<_i$
(next step on dimension)  indicates the direct successor on the dimension  $i$; $\lhd_i$ (after some
steps on dimension) encodes the transitive closure of $<_i$; and $\bigcirc^n_i$ (exactly after $n$ steps on dimension $i$) calculates the $n$-th direct successor.

Hence, a \textit{multi-dimensional relational pattern} may be defined as a set of Datalog atoms, regarding $n$
 dimensions, in which there are non-dimensional atoms and each event may be related to another event
 by means of the operators $<_i$, $\lhd_i$ and $\bigcirc^n_i$, $1 \leq i  \leq n$. 

The background knowledge  $\mathcal B$ contains the definitions of  the operators $\bigcirc^k_i$ and
$\lhd_i$ used to prove the dimensional operators appearing in the patterns. 
 Given $S$  a multi-dimensional relational sequence, in  the following we will  indicate by $\Sigma$
 the set of Datalog clauses $\mathcal B\cup U$, where $U$ is the set of ground atoms in $S$. 
In order to calculate the frequency of a pattern over a sequence it is important to define the concept of sequence subsumption.
\begin{definition}[Subsumption]
Given $\Sigma  = \mathcal  B\cup U$, where  $U$ is  the set of  atoms in  a sequence  $S$, and
$\mathcal B$ is a background knowledge. A pattern $P$ \emph{subsumes} the sequence $S$ ($P \subseteq
S$), iff there exists an SLD$_{\mathrm{OI}}$-deduction of $P$ from $\Sigma$.  
\end{definition}
An SLD$_{\mathrm{OI}}$-deduction is  an SLD-deduction under Object Identity~\cite{ferilli02aiiaa}.  In the Object
Identity framework, within a clause, terms that are denoted with different symbols must be distinct,
i.e. they must represent different objects of the domain. 

\subsection{Feature Construction via pattern mining}
The first step of the \textsf{Lynx} system corresponds to a feature construction process obtained by
mining frequent patterns from sequences.
The   algorithm for frequent  multi-dimensional relational pattern  mining is based on  the same
idea  of the  generic  level-wise search  method, known  in  data mining  from the  \textsc{Apriori}
algorithm~\cite{agrawal96}. 
The  level-wise algorithm  makes a  breadth-first search  in the  lattice of  patterns ordered  by a
specialization relation  $\preceq$. The search  starts from the  most general patterns, and  at each
level of  the lattice the  algorithm generates  candidates by using  the lattice structure  and then
evaluates the frequencies  of the candidates. In  the generation phase, some patterns  are taken out
using  the monotonicity  of  pattern frequency  (if  a pattern  is  not frequent  then  none of  its
specializations is frequent). 

The generation of the  frequent patterns is based on a top-down  approach. The algorithm starts with
the most general patterns. 
Then, at  each step it tries to  specialise all the potential  frequent patterns, discarding
the non-frequent  patterns and storing the  ones whose length is  equal to the  user specified input
parameter  \emph{maxsize}.  Furthermore,  for  each  new refined  pattern,  semantically  equivalent
patterns are detected, by using the $\theta_{\mathrm{OI}}$-subsumption relation~\cite{ferilli02aiiaa}, and discarded. In  the
specialization  phase,  the  specialization  operator  under  $\theta_{\mathrm{OI}}$-subsumption  is
used. Basically, the operator adds atoms to the pattern. 

\subsubsection{The background knowledge}

The algorithm  uses a background knowledge  $\mathcal B$ (a  set of Datalog clauses)  containing the
sequence  and a  set  of constraints,  similar  to that  defined in SeqLog~\cite{lee04},  that must  be
satisfied by the generated patterns.  
In particular, some of the  constraint included in $\mathcal B$ are (see~\cite{esposito08fundamenta}
for more details):
\begin{itemize}
 \item \emph{maxsize(M)}: maximal pattern length;
 \item \emph{minfreq(m)}: this constraint indicates that the frequency of the patterns must be larger than $m$;
\item \emph{type(p) and mode(p)}: denote the type and the input/output mode of the predicate's arguments p, respectively. 
They are used to specify a language bias indicating which predicates can be used in the patterns and to formulate constraints on the binding of variables;
  \item \emph{negconstraint([$p_1, p_2,\ldots, p_n$])}: specifies a constraint that the patterns must not fulfill, i.e.  if the clause ($p_1, p_2,\ldots, p_n$) subsumes the pattern then it must be discarded;
 \item \emph{posconstraint([$p_1, p_2,\ldots, p_n$])}: specifies a constraint that the patterns must fulfill. It discards all the patterns that are not subsumed by the clause ($p_1, p_2,\ldots, p_n$);
 \item \emph{atmostone([$p_1, p_2,\ldots, p_n$])}: this constraint discards all the patterns that make true more than one predicate among  $p_1, p_2,\ldots, p_n$;
 \item \emph{key([$p_1,p_2,\ldots,p_n$])}: it is optional and specifies that each pattern must have one of the predicates $p_1, p_2, \ldots p_n$ as a starting literal.
\end{itemize}

\subsubsection{Frequency, Support and Confidence} Given a set of relational sequences $D$ defined over a set of classes $C$, then the \emph{frequency}
of a pattern $p$,  $\mathrm{freq}(p,D)$, corresponds to the number of sequences  $s \in D$ such that
$p$  subsumes  $s$. The  \emph{support}  of  a pattern  $p$  with  respect to  a  class  $c \in  C$,
$\mathrm{supp}_c(p,D)$ corresponds to  the number of sequences  $s \in D$ whose class  label is $c$.
Finally, the  \emph{confidence} of a  pattern $p$ with respect  to a class  $c \in C$ is  defined as
$\mathrm{conf}_c(p,D) = \mathrm{supp}_c(p,D) / \mathrm{freq}(p,D)$.

\subsubsection{The refinement step}
The refinement of patterns is obtained by  using a refinement operator $\rho$ that maps each pattern
to a set  of specialisations of the pattern, i.e.  $\rho(p) \subset \{p' | p  \preceq p'\}$ where $p
\preceq p'$ means that $p$ is more general  of $p'$ or that $p$ subsumes $p'$.  In particular, given
the set $\mathcal  D$ of dimensions, the set $\mathcal  F$ of fluent atoms, the  set $\mathcal P$ of
non-fluent atoms, for each $i \in \mathcal D$, the refinement operator for specialising the patterns
is defined as follows: 
\begin{description}
 \item \textbf{adding a non-dimensional atom}
 \begin{itemize}
  \item  the pattern $S$  is specialised  by adding  a non-dimensional  atom;
 \end{itemize}
 \item \textbf{adding a dimensional atom}
 \begin{itemize}
  \item the pattern $S$ is specialised by adding the dimensional atom $(x <_i y)$;
  \item the pattern $S$ is specialised by adding the dimensional atom $(x \lhd_i y)$;
  \item the pattern $S$ is specialised by adding the dimensional atom $(x \bigcirc^n_i y)$.
 \end{itemize}
\end{description}

The dimensional atoms are added if and only if there exists a fluent atom referring to its starting event. The
length of a pattern $P$ is equal to the number of non-dimensional atoms in $P$. 

For each specialisation  level, before to start the next refinement  step, \textsf{Lynx} records all
the obtained patterns.  Hence, it could happens to have in the final set a pattern $p$ that
subsumes  a lot  of other  patterns in  the  same set.  However, the  subsumed patterns  may have  a
different support, contributing in different way to the classification model.

\subsection{Pattern-based Classification}
After having identified the set of frequent patterns, now the task is how to use them as features in order to
correctly classify unseen sequences.  
Let $\mathcal X$ be the input space of relational sequences, and let $\mathcal Y = \{1,2,\ldots,Q\}$
denote the finite set of possible class labels. Given a training set $D = \{(X_i, Y_i)|1 \leq i \leq
m\}$, where  $X_i \in \mathcal X$ is  a single relational sequence  and $Y_i \in \mathcal  Y$ is the
label associated to $X_i$, the goal is to learn a function $h : \mathcal X \rightarrow \mathcal Y$
from $D$ that predicts the label for each unseen instance. 

Let $\mathcal P$, with  $|\mathcal P| =d$, be the set of constructed  features obtained in the first
step of the \textsf{Lynx} system (the patterns mined
from $D$). For  each sequence  $X_k \in  \mathcal  X$ we  can build  a $d$-component  vector-valued
$\mathbf x =  (x_1, x_2, \ldots, x_d)$  random variable where each $x_i  \in \mathbf x$ is  1 if the
pattern $p_i \in \mathcal P$ subsumes the sequence $x_k$, and 0 otherwise.  


Using the Bayes' theorem,  if $p(Y_j)$ describes the prior probability of  the class $Y_j$, then the
posterior   probability  $p(Y_j|   \mathbf   x)$  can   be   computed  from   $p(\mathbf  x|   Y_j)$
by 
\begin{equation}
p(Y_j|\mathbf x) = \frac{p(\mathbf x | Y_j)p(Y_j)}{\sum_{i=1}^Q p(\mathbf x|Y_i)p(Y_i)}.
\end{equation}
Given a set of discriminant functions $g_i(\mathbf x)$,  $i = 1, \ldots, Q$, a classifier is said to
assign the vector $\mathbf x$ to the class $Y_j$ if $g_j(\mathbf x) > g_i(\mathbf x)$ for all $j \neq i$. 
Taking $g_i(\mathbf x) = P(Y_i |  \mathbf x)$, the maximum discriminant function corresponds to
the  \emph{maximum a  posteriori}  (MAP) probability.  For  minimum error  rate classification,  the
following discriminant function will be used
\begin{equation}
g_i(\mathbf x) = \ln p(\mathbf x | Y_i) + \ln P(Y_i).
\label{eq:discr3}
\end{equation}

Here,  we are  considering  a  multi-class classification  problem  involving discrete  features,
multi-class  problem in which  the components  of the  vector $\mathbf  x$ are  binary-valued and
conditionally independent. In particular, let the component of the vector $\mathbf x = (x_1, \ldots,
x_d)$          be           binary          valued          (0           or          1).          We
define $$p_{ij}=\mathrm{Prob}(x_i=1|Y_j)_{\substack{i=1,\ldots,d\\j=1,\ldots,Q}}$$ 
with the components of $\mathbf x$ being statistically independent for all $x_i \in \mathbf x$.
In this  model  each  feature $x_i$ gives us  a yes/no answer  about the
pattern $p_i$. However, if $p_{ik} > p_{it}$ we expect the $i$-th pattern to subsume a sequence more
frequently when its class is $Y_k$ than when it is $Y_t$. 
The factors $p_{ij}$ can be estimated from the training examples as frequency counts, as follows
\begin{eqnarray}
  p_{ij} && =\mathrm{Prob}(x_i=1|Y_j)  \nonumber \\
 && = \textrm{support}_{Y_j}(p_i)_{\substack{i=1,\ldots,d\\j=1,\ldots,Q}}. \nonumber
\end{eqnarray}
In  this  way,  the constructed  features  $p_i$  may  be  viewed as  \emph{probabilistic  features}
expressing the relevance for the pattern $p_i$ in determining the classification $Y_j$.

By assuming  conditional independence we
can write $P(\mathbf x | Y_i)$ as a  product of the probabilities of the components of $\mathbf
x$.  Given  this  assumption,  a  particularly  convenient  way  of  writing  the  class-conditional
probabilities is as follows: 
\begin{equation}
 P(\mathbf x| Y_j) = \prod_{i=1}^d (p_{ij})^{x_i}(1 - p_{ij})^{1-x_i}
\end{equation}

Hence, the Equation~\ref{eq:discr3} yields the discriminant function
\begin{eqnarray}
g_j(\mathbf x) = && \ln p(\mathbf x | Y_j) + \ln p(Y_j)=\nonumber \\
&& \ln \prod_{i=1}^d (p_{ij})^{x_i}(1 - p_{ij})^{1-x_i} + \ln p(Y_j)=\nonumber \\
&& \sum_{i=1}^d \ln \left ((p_{ij})^{x_i}(1 - p_{ij})^{1-x_i} \right ) + \ln p(Y_j)=\nonumber \\
&& \sum_{i=1}^d x_i \ln \frac{ p_{ij}}{1 - p_{ij}} + \sum_{i=1}^d \ln (1 - p_{ij})  + \ln p(Y_j)
\label{df}
\end{eqnarray}
The factor  corresponding to the prior  probability for the class  $Y_j$ can be  estimated from the
training set as $$p(Y_i) = \frac{|\{(X,Y) \in D \ \textrm{s.t.} \ Y = Y_i\}|}{|D|}, 1 \leq i \leq Q.$$
The minimum probability  of error is achieved by the following  decision rule: decide $Y_k$
if $g_k(\mathbf x) \geq g_j(\mathbf x)$ for all $j$ and $k$, where $g_i(\cdot)$ is defined as in
Equation~\ref{df}. 
Let we note that this discriminant function is linear in the $x_i$ and thus we can write
\begin{equation}
 g_j(\mathbf x) = \sum_{i=1}^d \alpha_i x_i + \beta_0,
\end{equation}
where 
$ \alpha_i = \ln (p_{ij} / (1-p_{ij}))$, and $ \beta_0 = \sum_{i=1}^d \ln (1- p_{ij}) + \ln p(Y_j)$.
Recall that we decide $Y_i$ if $g_i(\mathbf x) \geq g_k(\mathbf x)$ for all $i$. 
The magnitude of the  weight $\alpha_i$ in $g_j(\mathbf x)$ indicates the  relevance of a subsumption
for  the  pattern  $p_i$  in  determining  the  classification  $Y_j$.  This  is  the  probabilistic
characteristic of  the features obtained in the  feature construction phase, opposed  to the Boolean
feature.

\subsection{Feature Selection with stochastic local search}
After having constructed  a set of features, and  presented a method to use those  features to classify
unseen sequences, now the problem is how to find an optimal subset of
these features that optimise the prediction accuracy.
The optimisation problem of selecting a subset of features (patterns) with a superior classification performance
may be formulated as follows. Let $\mathcal P$ be the constructed original set of patterns, and let $f :
2^{|\mathcal P|} \rightarrow  \mathbb R$ a function scoring a selected  subset $X \subseteq \mathcal
P$. The problem of 
feature selection is to find a subset $\widehat{X} \subseteq \mathcal P$ such that 
$$f(\widehat{X}) = \max_{Z \subseteq \mathcal P} f(Z).$$
An exhaustive approach to this problem would require examining all $2^{|\mathcal P|}$ possible subsets of the
feature set $\mathcal  P$, making it impractical for  even low values of $|\mathcal P|$.  The use of
a  stochastic local search  procedure allows  us to  obtain \emph{good}  solutions without  having to
explore the whole solution space. 


Given a subset $P \subseteq \mathcal P$, for each sequence $X_j \in \mathcal X$ we let the classifier finds
the MAP hypothesis adopting the discriminant function reported in Eq.~\ref{eq:discr3}: 
\begin{equation}
\widehat{h}_P( X_j) = \argmax_{i} g_i(\mathbf x_j),
\end{equation}
where $\mathbf  x_j$ is the  feature based
representation of the sequence $X_j$ obtained using the patterns $P$.
Hence the  initial optimisation problem corresponds  to minimise the  expectation $$\mathrm E[\mathbf
1_{\widehat{h}_P(X_i)  \neq   Y_i}]$$  where  $\mathbf   1_{\widehat{h}_P(X_i)  \neq  Y_i}$   is  the
characteristic function of the training example $X_i$, defined as 
$$
\mathbf 1_{\widehat{h}_P(X_i)  \neq   Y_i} = \left\{ \begin{array}{rl}
 1 &\mbox{ if $\widehat{h}_P(X_i) \neq Y_i$} \\
 0 &\mbox{ otherwise}
       \end{array} \right.
$$

Finally, given $D$ the  training set with $|D|=m$ and $P$ a  set of features (patterns), 
the number of classification errors made by the Bayesian model is
\begin{equation}
err_D(P) = m \mathrm E[\mathbf 1_{\widehat{h}_P(X_i) \neq Y_i}].
\label{eq:eval}
\end{equation}

\subsubsection{GRASP$^{\mathrm{FS}}$}
\label{sec:grasp}

Consider  a \emph{combinatorial optimisation}  problem, where  one is  given a  discrete set  $X$ of
solutions and an objective function $f : X \rightarrow \mathbb R$ to be minimised and seeks
a solution $x^* \in X$ such that $\forall x \in X: f(x^*) \leq f(x)$. 
A  method to  find  high-quality solutions  for  a combinatorial  problem is  a  two steps  approach
consisting    of   a    greedy   construction    phase    followed   by    a   perturbative    local
search~\cite{sls04}.  The greedy  construction  method starts  the  process  from an  empty
candidate solution  and at  each construction  step adds the  best ranked  component according  to a
heuristic  selection  function. Then,  a  perturbative local  search  algorithm,  searching a  local
\emph{neighborhood}, is  used to improve the  candidate solution thus obtained.   Advantages of this
search method are the much better solution quality and fewer perturbative improvement steps to reach
the local optimum.  

Greedy Randomised Adaptive  Search Procedures (GRASP)~\cite{feo95} solve the  problem of the limited
number  of different  candidate  solutions generated  by  a greedy  construction  search method  by
randomising the  construction method.  GRASP is an  iterative process combining at  each iteration a
construction and a local  search phase. In the construction phase a  feasible solution is built, and
then its neighbourhood is explored by the local search. 

Algorithm~\ref{alg:grasp} reports the GRASP$^{\mathrm{FS}}$ procedure included in the \textsf{Lynx}
system to  perform  the feature
selection task.  In each iteration, it computes a  solution $S \in \mathcal S$ by using a randomised
constructive search procedure and then applies a  local search procedure to $S$ yielding an improved
solution. The main procedure  is made up of two components: a constructive  phase and a local search
phase.  

\begin{algorithm}
\caption{\textsc{GRASP$^{\mathrm{FS}}$}}
\label{alg:grasp}
\begin{algorithmic}
\REQUIRE{$D$: the training set; $\mathcal P$:  a set of patterns (features); \emph{maxiter}: maximum
  number of iterations; $err_D(P)$: the evaluation function (see Eq.~\ref{eq:eval})}
\ENSURE{solution $\widehat{S} \subseteq \mathcal P$}
\STATE $\widehat{S} = \emptyset$, $err_D(\widehat{S}) = +\infty$
\STATE iter $= 0$
\WHILE{iter $<$ maxiter}
  \STATE $\alpha = $ rand(0,1)
  \STATE \emph{/* construction */}
  \STATE $S = \emptyset$; $i = 0$
  \WHILE{$i < n$}
    \STATE $\mathcal S = \{ S' | S' = add(S,A)\}$
    \STATE $\overline{s} = \max \{err_D(T) | T \in \mathcal S\}$
    \STATE $\underline{s} = \min \{err_D(T) | T \in \mathcal S\}$
    \STATE RCL $= \{S' \in \mathcal S | err_D(S') \leq \underline{s} + \alpha (\overline{s} - \underline{s})\}$
    \STATE select the new $S$, at random, from RCL
    \STATE $i \leftarrow i + 1$
  \ENDWHILE
  \STATE \emph{/* local search */}
  \STATE $\mathcal N = \{ S' \in neigh(S) | err_D(S') < err_D(S)\}$
  \WHILE{$\mathcal N \neq \emptyset$}
    \STATE select $S \in \mathcal N$
    \STATE $\mathcal N \leftarrow \{ S' \in neigh(S) | err_D(S') < err_D(S)\}$
  \ENDWHILE
  \IF{$err_D(S) < err_D(\widehat{S})$}
  \STATE $\widehat{S} = S$
  \ENDIF
  \STATE iter = iter + 1
\ENDWHILE
\STATE \textbf{return} $\widehat{S}$
\end{algorithmic}
\end{algorithm}

The  constructive  search  algorithm  used  in GRASP$^{\mathrm{FS}}$  iteratively  adds  a  solution
component  by  randomly selecting  it,  according  to a  uniform  distribution,  from  a set,  named
\emph{restricted  candidate list}  (RCL), of  highly ranked  solution components  with respect  to a
greedy  function  $g  :  \mathcal  S   \rightarrow  \mathbb  R$.   The  probabilistic  component  of
GRASP$^{\mathrm{FS}}$ is characterised  by randomly choosing one of the best  candidates in the RCL.
In our case the greedy function $g$ corresponds to the error function $err_D(P)$ previously reported
in Eq.~\ref{eq:eval}.
In particular, given $err_D(P)$, the heuristic function, and $\mathcal S$, the set of feasible
solutions, $$\underline{s} = \min \{ err_D(S) | S \in \mathcal S\}$$ and $$\overline{s} = \max \{
err_D(S)  | S  \in \mathcal  S\}$$ are  computed. Then  the RCL  is defined  by including  in it  all the
components $S$  such that  $$err_D(S) \geq  \underline{s} + \alpha  (\overline{s} -  \underline{s}).$$ The
parameter  $\alpha$  controls the  amounts  of  greediness and  randomness.  A  value  $\alpha =  1$
corresponds to a greedy construction procedure, while $\alpha = 0$ produces a random construction. 
As  reported  in~\cite{Mockus97},  GRASP  with  a  fixed  nonzero  RCL  parameter  $\alpha$  is  not
asymptotically convergent  to a global optimum.   The solution to make  the algorithm asymptotically
globally convergent, could be to randomly  select the parameter value from the continuous interval
$[0, 1]$ at the beginning of each iteration  and using this value during the entire iteration, as we
implemented in GRASP$^{\mathrm{FS}}$.
Hence, starting from  the empty set, in the  first iteration all the subsets  containing exactly one
pattern are considered  and the best is  selected for further specialisation. At  the iteration $i$,
the working  set of patterns  $S$ is  refined by trying  to add a  pattern belonging to  $\mathcal P
\setminus S$.

To improve  the solution generated by  the construction phase, a  local search is used.  It works by
iteratively replacing the current solution with a better solution taken from the neighbourhood of the
current solution while there is a better solution in the neighbourhood.
In order  to build  the neighbourhood of  a solution $S$, \emph{neigh(S)},  the following
operators have been used. Given $\mathcal P$ the set of patterns, and $S = \{p_1, p_2, \ldots, p_t\}
\subseteq \mathcal P$ a solution: 
\begin{description}
\item[\textbf{add}:] $S \rightarrow S \cup \{p_i\}$ where $p_i \in \mathcal P \setminus S$;
\item[\textbf{remove}:] $S \rightarrow S \setminus \{p_i\} \cup \{p_k\}$ where
  $p_i \in S$ and $p_k \in \mathcal P \setminus S$.
\end{description}

In particular, given a  solution $S \in \mathcal S$, the elements  of the neighborhood $neigh(S)$ of
$S$ are those solutions that can be  obtained by applying an elementary modification (add or remove)
to $S$. Local search starts from an  initial solution $S^0 \in \mathcal S$ and iteratively generates
a series of improving solutions $S^1, S^2, \ldots$.  At the $k$-th 
iteration,  $neigh(S^k)$  is searched  for  an  improving solution  $S^{k+1}$  such  that $err_D(S^{k+1})  <
err_D(S^{k})$. If  such a solution is  found, it is made  the current solution. Otherwise,  the search ends
with $S^{k}$ as a local optimum. 

\section{Experiments}

Experiments have  been conducted  on protein  fold classification, an  important problem  in biology
since  the  functions  of  proteins  depend  on  how  they  fold  up.  The  dataset,  already  used
in~\cite{KerstingG04,Kersting06,Gutmann06}  is  made  up  of  logical  sequences  of  the  secondary
structure of protein  domains. The task is to predict  one of the five most  populated SCOP folds of
alpha  and beta  proteins (a/b):  TIM beta/alpha-barrel  (c1), NAD(P)-binding  Rossmann-fold domains
(c2), Ribosomal protein L4 (c23), Cysteine hydrolase (c37), and Phosphotyrosine protein phosphatases
I-like  (c55). The  class of  a/b proteins  consists of  proteins with  mainly parallel  beta sheets
(beta-alpha-beta units). Overall, the class distribution is  721 sequences for the class c1, 360 for
c2, 274 for c23, 441 for c37 and 290 for c55. 

As in~\cite{Gutmann06}, we used a round robin approach~\cite{furnkranz02}, treating each pair of classes as a
separate  classification problem,  and the  overall  classification of  an example  instance is  the
majority vote among all pairwise classification problems.

\begin{table}
\centering
\begin{tabular*}{\textwidth}{@{\extracolsep{\fill}}|c|c|cccccccccc|c|}
\hline 
\multirow{2}{*}{Conf.} & \multirow{2}{*}{\textsf{Lynx}} & \multicolumn{10}{c|}{Folds} & \multirow{2}{*}{Mean}\\
& & 1 & 2 & 3 & 4 & 5 & 6 & 7 & 8 & 9 & 10 & \\ \hline
\multirow{2}{*}{0.95} & w/o GRASP$^{\mathrm{FS}}$ & 0.84 & 0.88 & 0.83 & 0.83 & 0.85 & 0.76 & 0.85 & 0.81 & 0.82 & 0.80 &
0.826\\
& w GRASP$^{\mathrm{FS}}$ & 0.88 & 0.92 & 0.88 & 0.88 & 0.89 & 0.84 & 0.93 & 0,87 & 0.90 & 0.93 & \textbf{0.878} \\
\hline 
\multirow{2}{*}{1.0} & w/o GRASP$^{\mathrm{FS}}$ & 0.89 & 0.94 & 0.84 & 0.92 & 0.94 & 0.88 & 0.91 & 0.89 & 0.88 & 0.87 &
0.896\\
& w GRASP$^{\mathrm{FS}}$ & 0.94 & 0.97 & 0.93 & 0.95 & 0.95 & 0.93 & 0.93 & 0.97 & 0.90 & 0.94 &
\textbf{0.942}\\
\hline 
\end{tabular*}
\caption{Cross-validated accuracy of \textsf{Lynx} with  and without feature selection on two values
  for the confidence.}
\label{tab:lynx}
\end{table}

Table~\ref{tab:lynx}  reports the  experimental results  of  a 10-fold  cross-validated accuracy  of
\textsf{Lynx}. Two experiments have been conducted, one with a confidence level equal to $0.95$ and the other
with  a confidence level  of $1.0$.  In particular,  given the  training data  $D$, we  imposed that
$\mathrm{conf}_c(p,D) = 0.95$ (resp. $\mathrm{conf}_c(p,D) = 1$).
For each experiment, \textsf{Lynx} has been applied  on the same data with
and without  feature selection. In particular, we  applied the classification on  the test instances
without applying GRASP$^{\mathrm{FS}}$ in order to have a baseline accuracy value. Indeed, as we can
see, the accuracy  grows when GRASP$^{\mathrm{FS}}$ optimises the feature  set, proving the validity
of the method adopted for the feature  selection task. Furthermore, the accuracy level grows up when
we  mine patterns  with  a  confidence level  equal  to $1.0$  corresponding  to save  \emph{jumping
emerging patterns}\footnote{A  jumping emerging  pattern is  a pattern with  non-zero support  on a
class an a  zero support on all  the other classes, i.e. with  a confidence equal to  1.} only. This
proves that jumping patterns have a discriminative power greater than 
\emph{emerging patterns}\footnote{An emerging pattern is a pattern with a grow rate greater that 1.} (when the confidence level is equal to $0.95$).

As as second experiment  we compared \textsf{Lynx} on the same data  to other statistical relational learning
systems,  whose  cross-validated  accuracies  are  summarised  in  Table~\ref{tab:comparisons}.   In
particular,  LoHHMs~\cite{Kersting06}   were  able  to  achieve   a  predictive  accuracy   of  $75$\%,  Fisher
kernels~\cite{KerstingG04} achieved an  accuracy of about $84$\%, TildeCRF~\cite{Gutmann06}  reaches an accuracy value
of $92.96$\%, while \textsf{Lynx} obtains an accuracy of $94.15$\%. Hence, we can conclude that \textsf{Lynx} performs
better than established methods on real-world data.

\begin{table}
\centering
\begin{tabular}{|lc|}
\hline 
System & Accuracy\\
\hline 
LoHMMs~\cite{Kersting06} & 75\%\\
Fisher kernels~\cite{KerstingG04} & 84\%\\
TildeCRF~\cite{Gutmann06} & 92.96\%\\
\textsf{Lynx} & 94.15\%\\
\hline 
\end{tabular}
\caption{Cross-validated accuracy of LoHHMs, Fisher kernels, TildeCRF and \textsf{Lynx}}
\label{tab:comparisons}
\end{table}

\section{Conclusions}
\label{sec:conclusions}

In this paper  we considered the problem of multi-class relational  sequence learning using relevant
patterns discovered  from a  set of labelled  sequences. We  firstly applied a  feature construction
method in  order to map each  relational sequence into a  feature vector. Then,  a feature selection
algorithm to find an optimal subset of the constructed features leading to high classification
accuracy has been applied. 
The feature selection task has been solved  adopting a wrapper approach that uses a stochastic local
search algorithm embedding a na\"ive Bayes classifier. The performance of the proposed method applied
to a real-world dataset shows an improvement when compared to other
established methods.



\bibliographystyle{splncs}
\bibliography{biblio}

\end{document}